\documentclass[letterpaper, 10 pt, conference]{ieeeconf}  

\IEEEoverridecommandlockouts                              

\usepackage{svg}
\usepackage{amsmath}
\usepackage{amsfonts}
\usepackage{booktabs,siunitx}

\usepackage{graphicx}
\usepackage{comment}
\usepackage[caption=false]{subfig}
\let\labelindent\relax
\usepackage{enumitem}
\usepackage{todonotes}
\usepackage{url}

\title{\LARGE \bf
A Multimodal Data Set of Human Handovers with Design Implications for Human-Robot Handovers}

\author{Parag Khanna$^{1}$, Mårten Björkman$^{1}$ and Christian Smith$^{1}$
\thanks{$^{1}$ Division of Robotics, Perception and Learning (RPL), EECS, KTH Royal Institute of Technology, Sweden
        {\tt\small paragk@kth.se,} {\tt\small celle@kth.se,} {\tt\small ccs@kth.se}}%
}

\usepackage{hyphenat}
\begin{document}

\maketitle
\thispagestyle{empty}
\pagestyle{empty}

\begin{abstract}
Handovers are basic yet sophisticated motor tasks performed seamlessly by humans. They are among the most common activities in our daily lives and social environments. This makes mastering the art of handovers critical for a social and collaborative robot. In this work, we present an experimental study that involved human-human handovers by 13 pairs, i.e., 26 participants. We record and explore multiple features of handovers amongst humans aimed at inspiring handovers amongst humans and robots. With this work, we further create and publish a novel data set of 8672 handovers, bringing together human motion and the forces involved. We further analyze the effect of object weight and the role of visual sensory input in human-human handovers, as well as possible design implications for robots. As a proof of concept, the data set was used for creating a human-inspired data-driven strategy for robotic grip release in handovers, which was demonstrated to result in better robot to human handovers.
\end{abstract}

\section{Introduction}

Robots have been used to increase productivity and carry out risky jobs in industrial settings with success. Such industrial robots are technically advanced and suited for work in isolation on particular tasks. However, significant recent advancements in robotic intelligence and technology have made it possible to foresee robots cohabiting with humans to help and collaborate. We envisage social robots having a shared environment with humans performing a variety of activities in the future. These activities include, but are not limited to, assisting at home \cite{Robotic_assistants_in_personal_care_BILYEA20171}, collaborating in factories \cite{Safe_Human_Robot_Collaboration_in_Industrial_Robla_8107677}, helping humans with restricted mobility and in old age homes \cite{Robots_for_Elderly_Care_Bardaro}. Indeed, effective human robot collaboration (HRC) will also allow humans and robots to complement each other's abilities. In a real world environment filled with uncertainties, the decision making skills of a human counterpart can largely benefit a collaborative robot. This has motivated many robotic platforms aimed at physical HRC \cite{physical_HRC_robots_survey8907351}.
While doing collaborative work, robots must be able to engage with people in a safe, natural, and convenient manner. Thus, it is crucial that we create robots that can interact with people physically. Handover is one such physically interactive task which frequents our daily social lives. 
To achieve effective human robot collaboration, fluent and natural robot human handovers are
essential. As a result, a social robot needs to master the skill of handover. 

\begin{figure}[t]
      \centering
     \includegraphics[width=6.3cm,height=6.3cm,trim={5.5cm 0.4cm 5.5cm 0.5cm},clip]{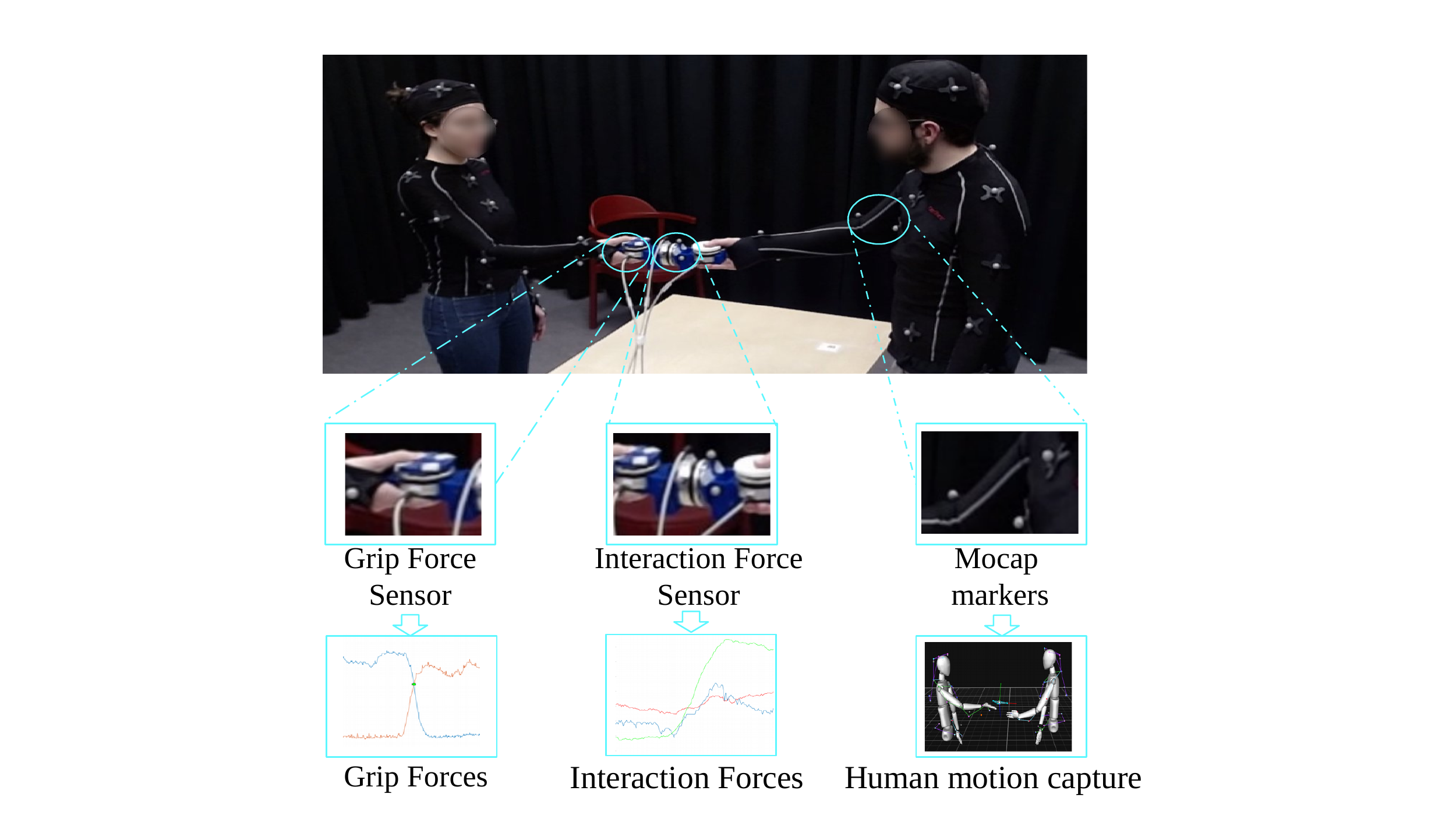}  
       \setlength\abovecaptionskip{-0.2\baselineskip}
      \caption{A snippet of the human study exhibiting various aspects of a handover and the recorded signals}
      \label{fig:H2H_handover_study}
\end{figure}
According to many studies \cite{chan_grip_from_load_second_PR2,Modelling_human_reaching_phase_inH2H_forR2H_sina,When_where_how_human-human_studyStrabala}, humans prefer robot behaviour that mimics humans. Humans also make the ideal subjects for studying handovers due to their proficiency and adaptation skills. In this paper, we present an experimental study of human handovers, which is used to create an open source data set of handovers and to further analyze various aspects of human handovers as shown in Fig. \ref{fig:H2H_handover_study}. We also focused on studying the effect of increased weight of transfer in handovers. We analyze handovers and what happens when humans rely on just haptic feedback, comparing the cases with no visual information to those with visual and haptic feedback. The aim of this work is to learn from human handovers to inspire natural, safe, and efficient handovers between humans and robots. 
The main contribution of this work is a novel multimodal data set of 8672 human-human handovers with information from multiple sensors including grip forces, interaction force and human motion tracking.
We also provide some qualitative observations including the effects of increased weight of transfer and visual sensory impairment on various aspects of handovers aimed to enhance handover capabilities for robots.

\section{Background and Related Work}

A task of handover is comprised of multiple phases that humans collaborate to perform seamlessly. Verbal or nonverbal communication is used to initiate a coordinated spatio-temportal movement of a giver and a taker. In case of a general handover, the giver determines a suitable location to transfer the object in the inter-personal space shared with the taker and starts moving the object towards that location \cite{human_human_to_human_robot_study_Controzzi} \cite{survey_review_2022_object_handovers}. The taker, too, prejudges the handover location based on the giver's motion with the object and starts approaching the location with the preferred hand. As handover progresses, both giver and taker adjust their arm speed based on visual sensory information. The coordinated movement finally ends with a light impact on the object as the taker starts interacting with the object, trying to form a grip on it. During this interaction, the giver decreases the grip force on the object while the taker's grip force increases. Both giver and taker also share the responsibility of sharing the object's weight and maintaining the object's pose so that it does not fall in the process. The handover ends when the giver releases the object completely after determining a stable grasp on the object by the taker. Thus, a robot should be competent across multiple phases of a handover for fluent handovers.

Furthermore, there are more complex handover scenarios to which humans swiftly adapt. Humans are not bothered by different-weighted objects from their daily lives, for example, handing over a knife and a hammer with fluency \cite{survey_review_2022_object_handovers, When_where_how_human-human_studyStrabala, object_orientation_dataset_chan}. There are also common scenarios when the weight of the object to be transferred turns out to be different than expected. One such scenario is the handover of a supposedly empty box, which turned out to be full. Both human giver and taker are capable of quickly adapting to changes caused due to these uncertainties regarding the weight of the transferred object. In some scenarios, a human giver does not look actively at the object, or does not look at all, while giving the object. In such a case, the giver relies on only the haptic feedback to perform the handover, i.e., sensing the forces in interaction. In these scenarios too, both giver and taker are able to suitably adapt and perform a safe and efficient handover. A social robot is likely to face these scenarios, owing to the dynamic social environment. It is expected to adapt and perform according to the given situation in order to accommodate safe handovers, while avoiding failures. 

The interest in human inspired handovers for robots has led to multiple studies on handovers amongst humans \cite{survey_review_2022_object_handovers}. 
Several human studies have been done to study human motion to model the movement of a robotic arm close to a human in handovers. Human hand motion was captured in human handovers by an electromagnetic tracker and markers placed on the hand in \cite{robot_reaching_profiles_4600651}, and was further used to evaluate robot reaching profiles for handovers. Human arm motions were recorded via a camera in \cite{human-like-motion-forR2H_Handovers_Rasch2018AJM} to analyse and propose a joint motion model for a robotic giver for human like handovers. To understand verbal and non-verbal human cues, and joint coordination involved in human handovers, a human study was done in \cite{When_where_how_human-human_studyStrabala} in a kitchen environment with multiple color and depth cameras for analysis.

To study forces in human handovers, \cite{chan_grip_from_load_second_PR2} used a baton with grip force sensors and a load force sensor for handovers in a human study. 
The baton was vertically transferred in handovers 
and a relationship between grip forces and load forces was measured, which was used to develop a controller for robotic grip forces given load forces.
In \cite{Effect_of_vision_speed_on_Grip_Dohring2020}, the effect of reaching velocities and sensory manipulations was seen on grip forces in human human handovers. In one of the studied cases, the giver's vision was manipulated by closing the eyes during the handover. It was observed that sensory manipulation had a strong effect on temporal aspects of the handover. The effect of object weight on handover location and duration was investigated by a human study in \cite{h2H_handovers_how_dis_and_obj_mass_matter_Clint2017}. It was shown that only the handover duration was impacted by object weight. In \cite{human_human_to_human_robot_study_Controzzi}, a relation between the taker arm reaching speed and the giver's grip forces was found in a human study.
The effect of restricting visual information on giver was also studied to conclude that coordinated arm-movement is impacted by lack of visual feedback.

Some publicly available data sets do exist for human handover tasks. The \cite{dataset_Emaro_CARFI2019109} data set includes human motion tracking via different sensors for participants in human handovers passing multiple objects. They provide 3D upper skeleton tracking via a Motion-Capture (MoCap) system and a depth camera, as well as inertial data of the hands from smartwatches. In \cite{object_orientation_dataset_chan}, a human study was done to inspire proper object orientations in robot to human handovers. The corresponding data set includes MoCap tracking data of the upper skeleton and object pose in handovers. 
Human grasps were studied and labelled in \cite{more_precision_type_grasp_type_location_study_dataset_cini_controzzi}, collecting over 5202 grasps. Their study showed that precision grasp is the most preferred one. Also most givers in human handovers preferred a precision grasp irrespective of object. 
In \cite{h2o_dataset}, an image based data set focused on hand poses and grasps via visual analysis of handovers is described, where the hand poses are tracked by markers placed on each finger. 

We observe that the task of handover has multiple aspects and modalities spanning across spatio-temporal and force domains. Many of these aspects have been researched, but independently from each other. Publicly available data sets also lack information about grip and interaction forces. 
In this work, we present a comprehensive study that integrates multiple aspects of handovers and a multimodal data set containing handover forces with human motion tracking. In a multimodal interaction like handover, it is important to understand not only the effect of each modality, but also how different modalities are interrelated. Moreover, some of these modalities, like human-grip force, might not be available to a robot in handovers. However, a multimodal dataset makes it possible to train the robot on multiple modalities and to learn mapping from one to another. The robot can then use only those modalities that are accessible during online execution.


\section{Human Handovers Study}
Humans practice handovers countless times over several years to perfect them. This motivated us to design an experimental study focused at human handovers. We aimed to get a unified perspective of the giver's and taker's motion, grip forces applied by them, and the interaction forces involved in human handovers. Our goal in this study is two-fold: 
\begin{enumerate}
    \item Recording and analysing \textit{natural} human handovers.
    \item Create a multi-sensor data set of human handovers.
\end{enumerate}

\subsection{Participants}
For the study we recruited a total of 26 participants which included 22 male and 4 female participants between the ages of 14 and 28 (age $23.23\pm 2.55$ years). Only one male participant was left handed while others were right handed. This lead to 13 participant pairs which comprised of one female-female pair, two female-male pairs and ten male-male pairs. One of these male-male pairs was a right handed-left handed pair. The participants had little to no prior acquaintance with each other. They were recruited by advertisement amongst the university students and were rewarded with a 100 SEK gift-voucher.

\subsection{Experimental Design}

\subsubsection{Baton}
\begin{figure}[t]

  \centering
  \setlength\abovecaptionskip{-0.25\baselineskip}
  \includegraphics[scale=0.35,trim={4cm 1.8cm 4.5cm 4.1cm},clip]{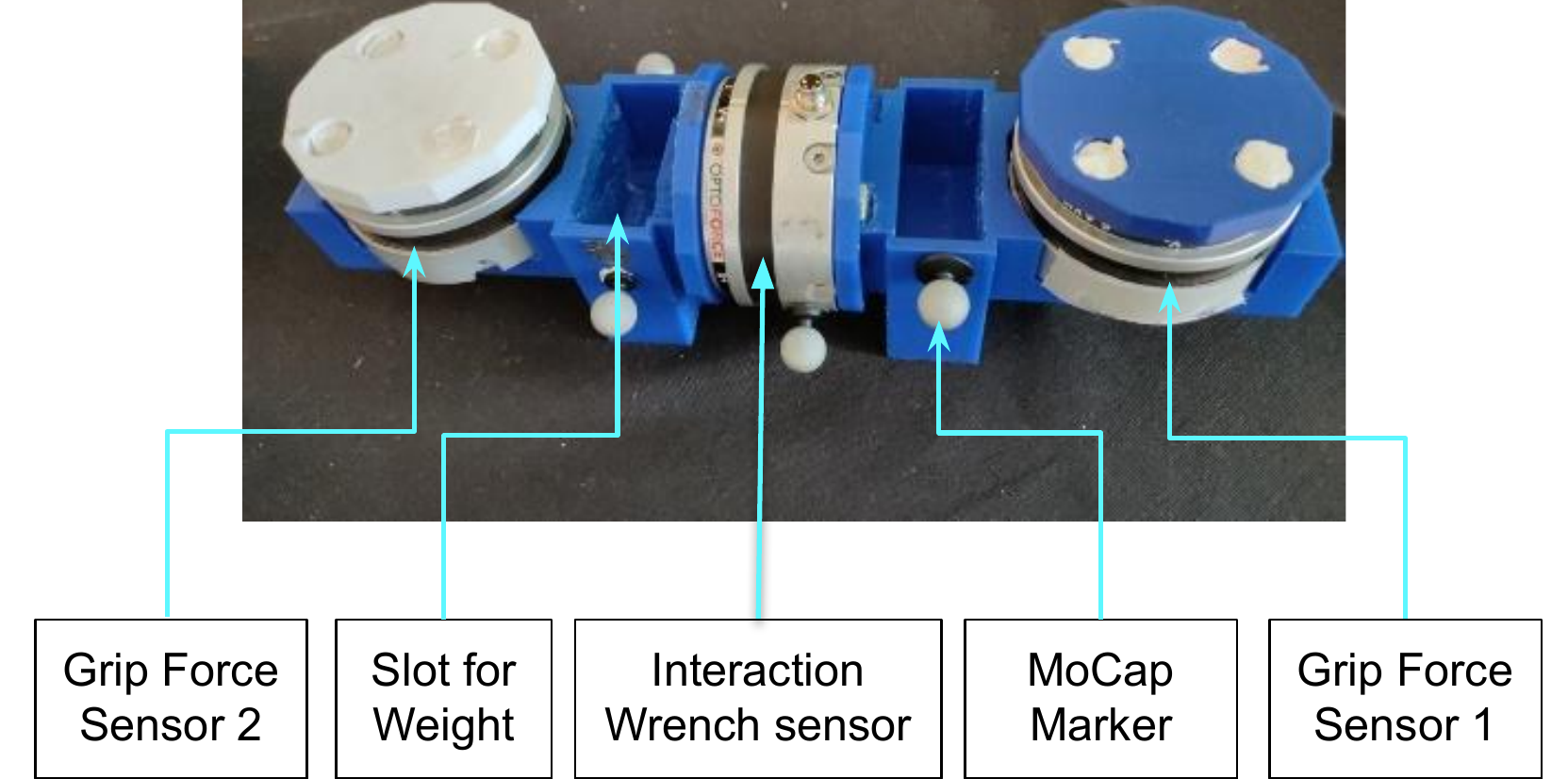}
  \caption{3D printed sensor embedded Baton}
  \label{fig:baton_details}
\end{figure}

The experiment involved passing back and forth a sensor embedded baton which was made by 3D printing. The baton houses three 6D Force/Torque (F/T) sensors from Onrobot \cite{Onrobots_F_T+Doe:2022:Online} as shown in Fig. \ref{fig:baton_details}, where each F/T sensor measures a 6D wrench, comprising 3 forces and 3 torques. The F/T sensor in the center of baton is used to measure the interaction wrench in the handover. The two sensors on the sides are used to measure the grip forces by the giver and the receiver. Additionally, 5 Motion Capture (MoCap) markers were placed asymmetrically on the baton to track its movement. The baton has a total weight of 0.8 kg and measures 0.28x0.085x0.075 m. It also has two slots to add external lead-weights, increasing the baton weight when needed. 
The top of the two sensors for grip force were color coded with blue and white parts. For simplicity, participant 1 was told to always pick up and receive the baton from the blue side while participant 2 would do so from white side. 

\begin{figure}[b]
  \centering
  \subfloat[]{
  \includegraphics[width=4.9cm,height=4cm,trim={5cm 4.4cm 4.5cm 3.3cm},clip]{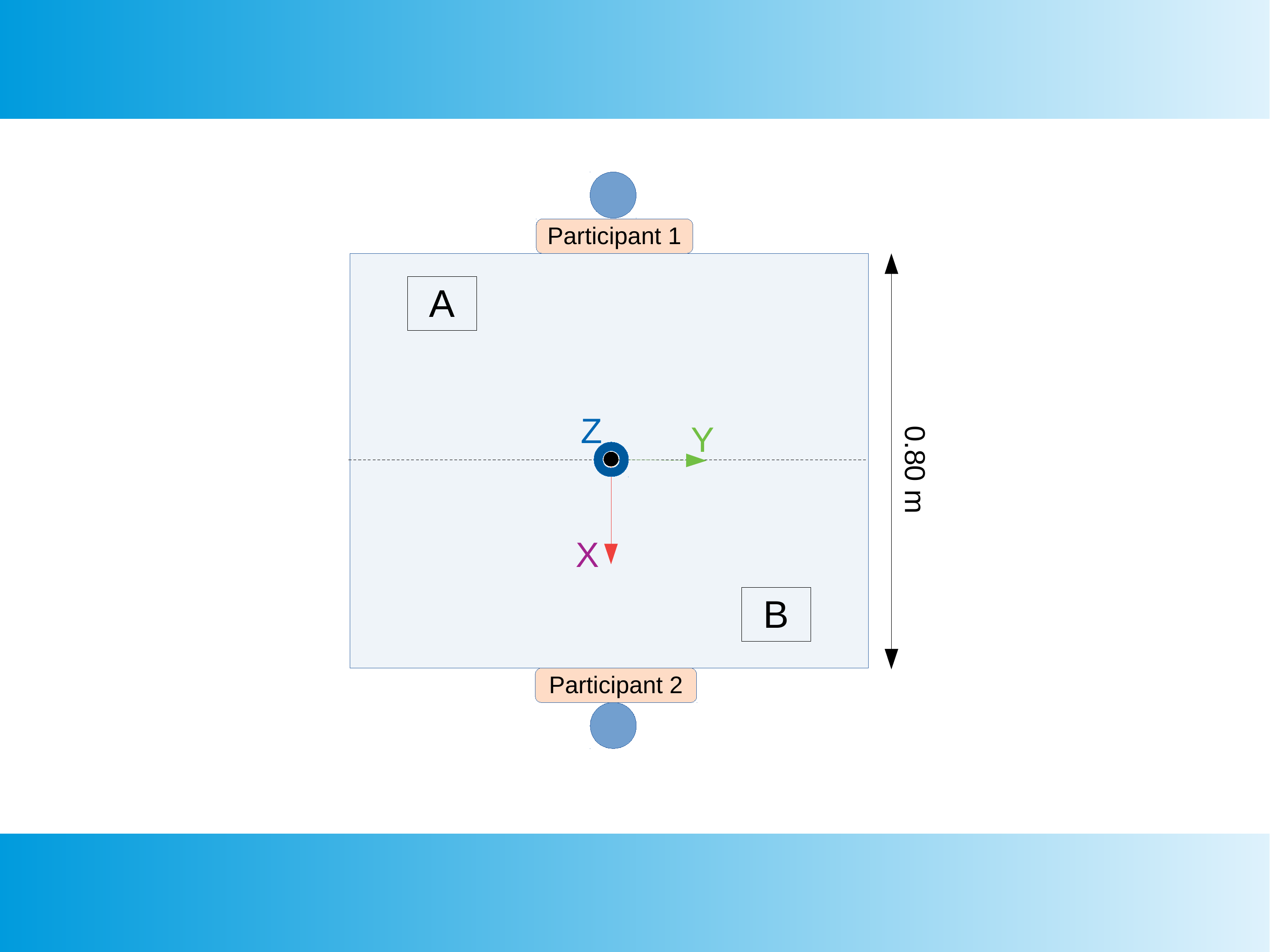}}
  \subfloat[]{
   \includegraphics[width=3.4cm,height=3.5cm,trim={4cm 4cm 4.5cm 8.0cm},clip]{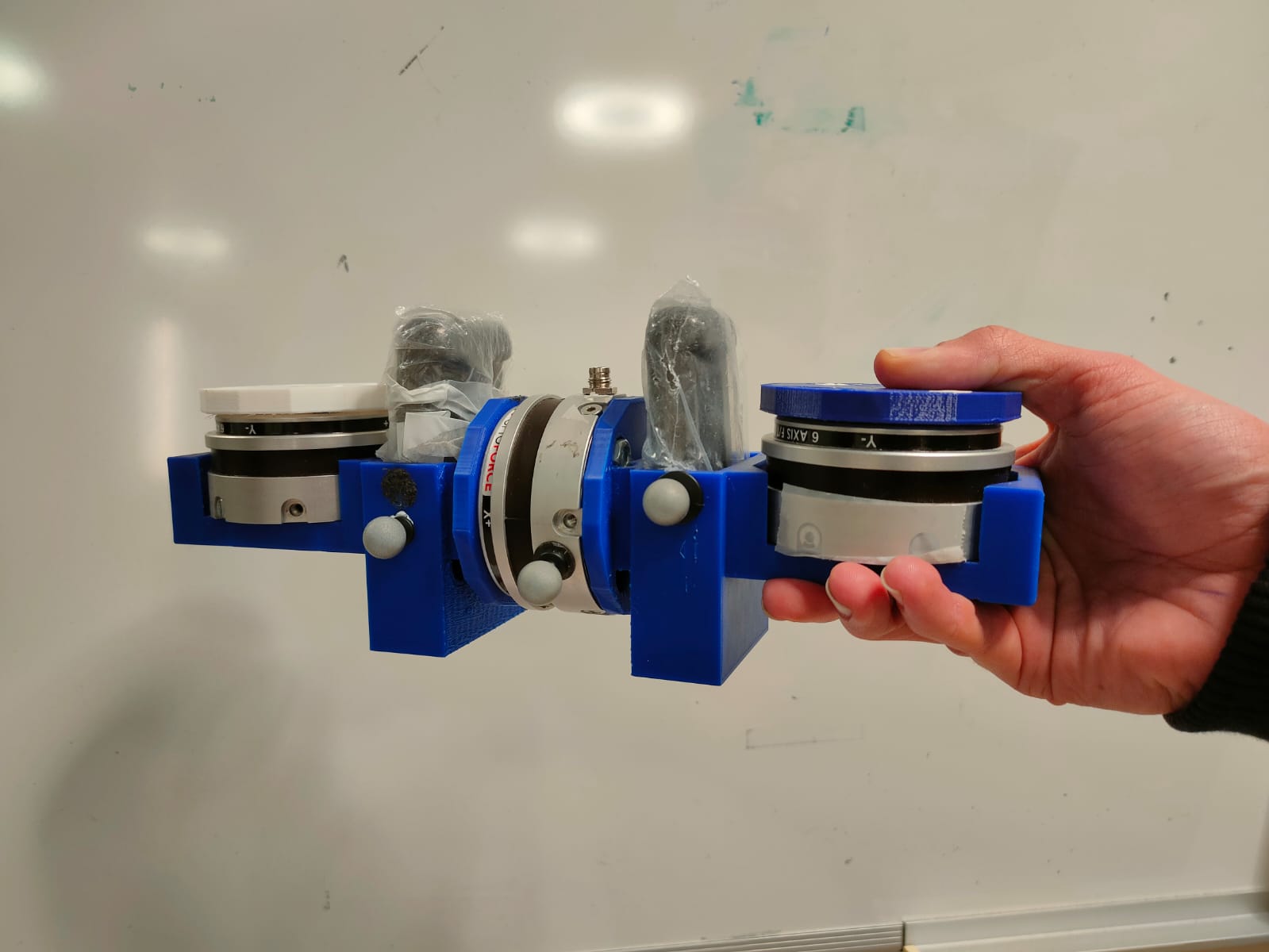}}
  \caption{ (a) Table (0.8 m wide) setup for human-human handovers (b) Precision grasp on baton with lead weights added}
  \label{fig:table_setup_and_precision_grasp}
\end{figure}
\subsubsection{Experimental Setup}
To record the movements of participants, the experiment was conducted in a MoCap room by Optitrack \cite{optitrack+Doe:2022:Online} and participants were required to wear upper body MoCap suits.
The participants performed handovers standing across a table as shown in Fig. \ref{fig:table_setup_and_precision_grasp}(a). The experimental session started with participant 1, acting as giver, picking up the baton from position A and handing over the baton mid-air and horizontally to participant 2, the receiver, who would then place the baton at position B. Further, participant 2 acting as giver, would pick up the baton and hand it over to participant 1, who places it back at position A, and the process would be repeated. To preserve the naturalness of handovers, they were not given any instruction regarding the speed and location of handover. To reduce ergonomic load, positions A and B were marked close to the dominant hand of participants. The layout in Fig. \ref{fig:table_setup_and_precision_grasp}(a) corresponds to two right handed participants.

\subsubsection{Experimental Settings}
The handovers occurred in three different settings, and the participants were required to do two sets of 9 minutes for each of the three settings:
\begin{itemize}[leftmargin=*] 
    \item Setting 1 (S1) - \textbf{Normal}: In this setting, the baton was used for handover without any external weight. Neither giver nor taker was visually impaired by closing the eyes. 
    The corresponding sets were named Set 1 and 2.
    \item Setting 2 (S2) - \textbf{Heavier Baton}: In this setting, two lead weights of 0.5 kg were added to the slots of the baton, leading to a total weight of 1.8 kg (Fig. \ref{fig:table_setup_and_precision_grasp}(b)). This enables us to analyze the effects of a significant increase in weight on different aspects of handovers, particularly the interaction forces. These sets are called Set 3 and 4.
    \item Setting 3 (S3) - \textbf{Giver vision impairment}: In this setting, the giver was asked to close the eyes while handing over the baton without added weights. The giver would pick the baton normally, with eyes closed  only during the giving phase as the receiver took the baton. The giver would open their eyes once the baton had been completely released. This scenario is similar to a robotic giver lacking vision input during handover and must rely solely on interaction forces. This setting allows us to analyze the effects of sensory impairment and the sets are called Set 5 and 6.
\end{itemize}

\subsection{Procedure}

\subsubsection{Introduction and Instructions}
Before each experimental session with a participant pair, they were given a brief outline of what they were required to do in the experiment. They were given verbal and written explanations of all steps, signed an informed consent form regarding the experiment.
They were also assigned as Participants 1 and 2. 
To ensure the naturalness of the handovers, they informed about the aim of the study only at the end of the experiment. They were suggested to get acquainted with each other and choose some mutual topics of interest to talk about while doing the handovers. This was an important step to reduce monotonicity in the experiment, which could lead to unnatural behavior.

The participants were given sanitized MoCap suits of their size to wear. After the MoCap markers were properly placed on their suits, they were allowed to enter the MoCap room and stand across the table. They were instructed to use precision grasp for holding the baton, which was then demonstrated as in Fig. \ref{fig:table_setup_and_precision_grasp}(b), and to do some practice handovers. A precision grasp has been chosen because it involves only two opposing grasp surfaces, as opposed to multiple oblique contact surfaces involved in other type of grasps, making it easier and more accurate to measure forces using a single sensor. This grasp type has also been used in similar studies for grip forces \cite{chan_grip_from_load_second_PR2, human_human_to_human_robot_study_Controzzi,loadsharing_pull_strategy-10.3389/frobt.2021.672995} and it is shown in \cite{grasp_types_in_house_and_machine_shop_zheng11,more_precision_type_grasp_type_location_study_dataset_cini_controzzi} that precision grips make up about 50$\%$ of all grips used in tool-intensive activities at homes and industry.
\subsubsection{Experimentation}
For all pairs, the experimentation started with Set 1 of handovers in normal setting. There was no restriction on how many handovers the participants need to do in the 9 minute allotted time for each set. The participants would then have a break of 2 minutes where they were allowed to sit in the chairs provided. The further order of sets were randomized so that no sets of similar strategies come after each other. After completing three sets of the experiment, they were given a longer break of minimum 5 minutes and the experiment continued when they were ready again. They were also allowed to take any breaks in between.
\begin{figure}[b]
     \centering
    \setlength\abovecaptionskip{-0.4\baselineskip}
    \includegraphics[width=150pt,height=4.2cm,trim={0.1cm 5.5cm 1.7cm 6cm},clip]{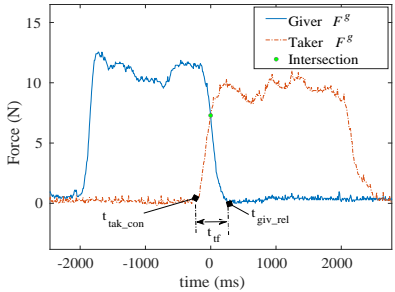}
      \caption{Grip force ($F^g$) variation in a particular handover.}
      \label{fig:forces_handover}
   \end{figure}

\subsection{Data Collection}
For each set of experimentation by a participant pair, multiple sensory data were recorded by reading through Robot Operating System (ROS) \cite{ROS-Doe:2022:Online} as common interface. The data from F/T sensors is read using publicly available libraries for ROS at 333Hz 
This data contains 6D wrench for the interaction and grip force sensors.
The tracking data from MoCap has the pose tracking of baton and upper body skeletal representation of the two participants which contains pose tracking of 13 segments: Hip, Ab, Chest, Neck, Head, Left-shoulder, Left-upper arm, Left-forearm, Left hand, Right-shoulder, Right-upper arm, Right-forearm, Right hand. Each segment frame of the skeleton is tracked with respect to the world frame, which lies on the floor beneath the center of the table. This data is broadcasted by MOTIVE software from Optitrack \cite{optitrack+Doe:2022:Online} at a frequency of 120Hz and read by a customized ROSnode. Furthermore, this ROSnode \cite{github_self_code} reads the original data from F/T sensors and republishes synchronized F/T and MoCap tracking data with common timestamps at 120Hz. 
For each body-segment, the tracking data is published in form of ROS-pose messages which contain 3D position and 4D quaternion rotation information. Finally, the recorded data for each experimental set involved the synchronized data from F/T sensors, the baton pose and skeletal representation for participant 1 and 2 at 120 Hz.

\section{Data set}
The continuous recordings of handovers for each pair was post-processed to separate individual handovers.
\subsection{Data set Creation}
\begin{figure}[t]
    \centering
    \setlength\abovecaptionskip{-0.04\baselineskip}
     \subfloat[]{\includegraphics[width=.45\linewidth,height=3cm,trim={1.1cm 6.5cm 1cm 7.35cm},clip]{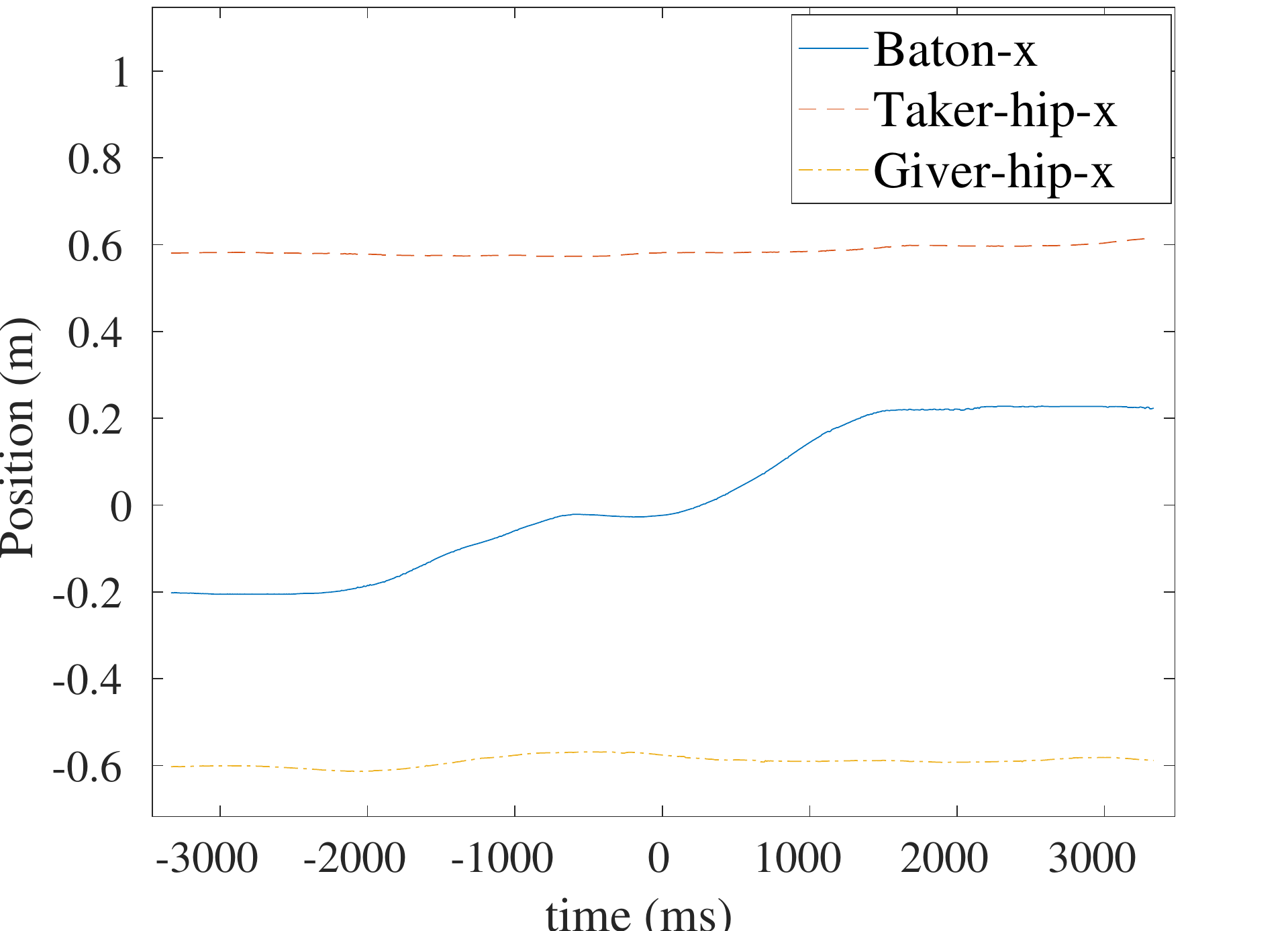}}
     \subfloat[]{\includegraphics[width=.45\linewidth,height=3cm,trim={0.8cm 6.5cm 1cm 7.45cm},clip]{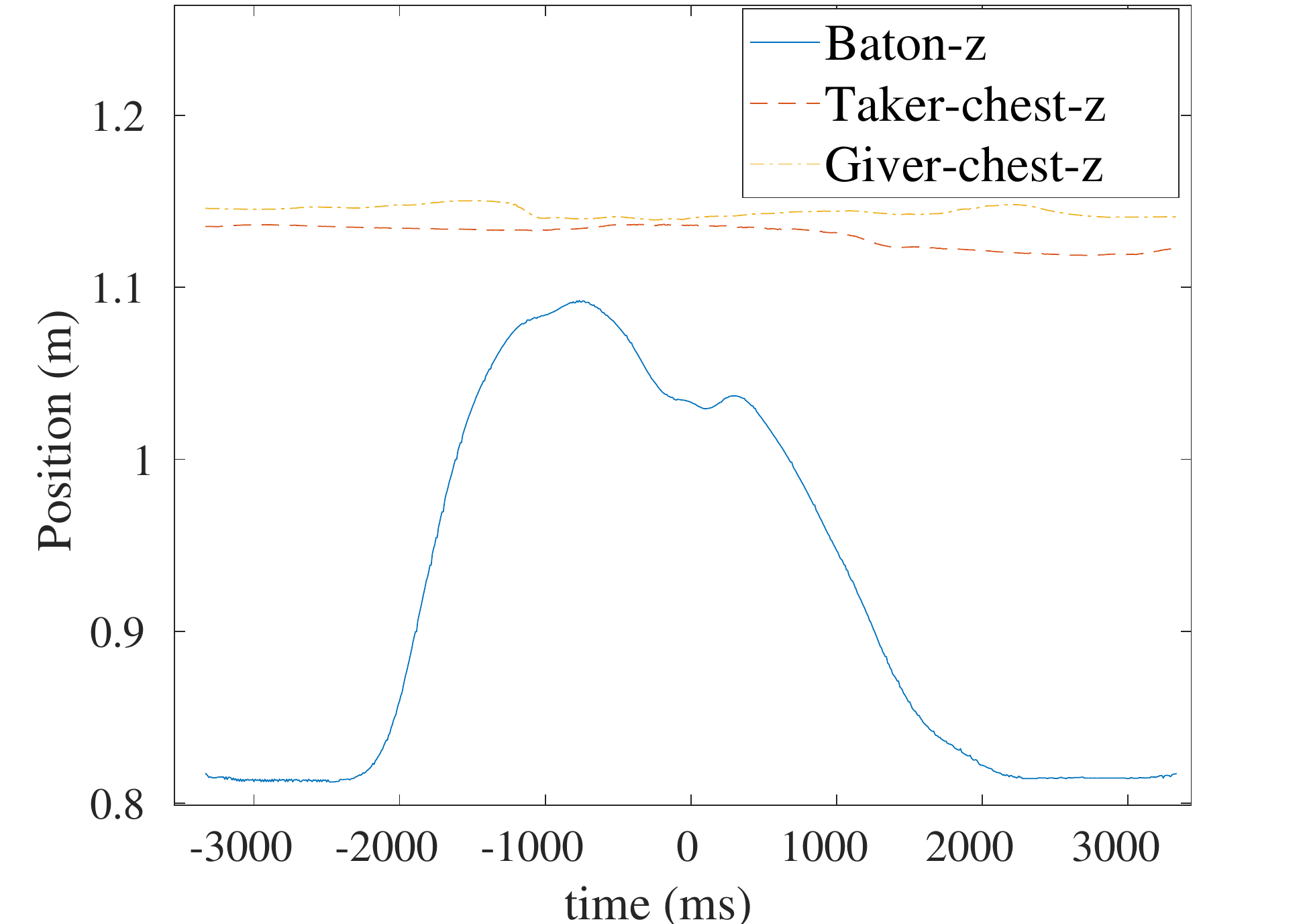}}
    \caption{Baton transfer location (a) Horizontal (b) Vertical}
    \label{fig:baton_location_handover}
\end{figure}
In every handover, the giver lets go of the object as it is transferred to the taker, who takes hold of it. As a result, for each handover, the giver's grip force decreases to zero, whereas the taker's grip force increases from zero, resulting in an intersection of grip forces. In force-space, a handover is shown in Fig. \ref{fig:forces_handover} with the intersection point of grip-forces ($F^g$) of giver and taker. The existence of this intersection point for each handover was leveraged to separate every handover in the recorded data. As can be seen in the Fig. \ref{fig:forces_handover}, the handovers are centered about the intersection point at $t=0$ ms. The separated handovers are saved for a duration of 3.333 seconds (400 timestamps at 120 Hz) before and after the intersection point. For all handovers, this duration was sufficient to capture the baton being picked up from the table by the giver, the handover to the taker and baton being placed again on the table.
The larger duration was used to verify that the baton was indeed transferred from the giver to taker as can be seen in Fig. \ref{fig:baton_location_handover}. It also enables analysis of multiple aspects of human motion before and after the baton transfer. For this data set, we saved the recorded data with common timestamps i.e. the republished data. The data set is publicly available at: 
\url{https://github.com/paragkhanna1/dataset}.
The data set contains a total of \textbf{8672} handovers, distributed among the 3 experimental settings as follows:
\begin{itemize}
    \item S1 - \textbf{2999} normal handovers.
    \item S2 - \textbf{2764} handovers with heavy baton.
    \item S3 - \textbf{2909} handovers with giver vision impairment. 
\end{itemize}

\subsection{Features}
\begin{table}[b]
\setlength\abovecaptionskip{-0.5\baselineskip}
  \caption{Signals of one recorded handover}
  \centering
  \label{tab:onehandover}
  \resizebox{0.8\columnwidth}{!}{
  \begin{tabular}{cccc}
    \toprule
    
    
    Signal & Signal & Signal  \\   
     No. & Name & Components \\
    
    \midrule
     1 & Wrench\_interaction & Force (x,y,z) \\
    & & Torque(x,y,z) \\
    2 & {Wrench\_giver}$^*$  & ---\texttt{"}--- \\
    3 & {Wrench\_taker}$^*$ & ---\texttt{"}---    \\
    4 & baton\_pose & Position(x,y,z)  \\
     & & {Rotation}($q_0,q_1,q_2,q_3$)\\
    5-17 & giver-Skeleton & 13 bodies-pose\\
    18-30 & taker-Skeleton & 13 bodies-pose\\
    \bottomrule
    \multicolumn{2}{c}{* Grip force is given by -Force(z) } 
    \end{tabular}
}
\end{table}
Table \ref{tab:onehandover} summarizes all information saved for an individual handover. It lists the signal names with which they are available in the data set. As stated earlier, each signal is centered around the grip-force intersection point and has a total duration of 6.675 seconds. The metadata includes the height, arm-lengths, age, and handedness of the two subjects.
\section{Data analysis}

\subsection{Transfer time - $t_{tf}$}

\begin{table}[t]
\vspace{1.5mm}
\setlength\abovecaptionskip{-0.5\baselineskip}
  \caption{Transfer time}
  \centering
  \label{tab:transfer time}
  \resizebox{0.9\columnwidth}{!}{
  
  \begin{tabular}{cccccc}
    \toprule
    
    
    Pair & \multicolumn{3}{c}{$t_{tf}$ (Mean $\pm$ SD, ms)} &     \multicolumn{2}{c}{\textit{t}-test significance}\\ \cmidrule(lr){2-4} \cmidrule(lr){5-6}
     No. & S1 & S2 & S3 & S1-S2 & S1-S3\\
    
    \midrule
    1 & \textbf{414.6 $\pm$127.0} & 481.4 $\pm$114.5 & 540.3 $\pm$160.8 & \textbf{\checkmark} & \textbf{\checkmark}\\
          
    2 & 545.5 $\pm$185.2 & 579.4 $\pm$163.3  & 601.3 $\pm$198.1 & \text{\sffamily X} & \text{\sffamily X}\\
          
    3 & \textbf{398.5 $\pm$136.4} & 484.7 $\pm$154.8  & 583.0 $\pm$198.5 & \textbf{\checkmark} & \textbf{\checkmark}\\
           
    4 & \textbf{471.9 $\pm$184.9} & 632.9 $\pm$181.5 & 560.7 $\pm$183.0 & \textbf{\checkmark} & \textbf{\checkmark}\\
           
    5 & \textbf{502.2 $\pm$169.4} & 610.9 $\pm$148.4 & 542.5 $\pm$166.6 & \textbf{\checkmark} & \textbf{\checkmark}\\
             
    6 & \textbf{475.5 $\pm$197.7} & 513.6 $\pm$174.7 & 660.9 $\pm$181.9 & \textbf{\checkmark} & \textbf{\checkmark}\\ 
            
    7 & \textbf{577.1 $\pm$168.0} & 622.6 $\pm$180.9 & 693.8 $\pm$176.2 & \textbf{\checkmark} & \textbf{\checkmark}\\
             
    8 & \textbf{474.3 $\pm$170.1} & 654.1 $\pm$166.1 & 649.1 $\pm$178.4 & \textbf{\checkmark} & \textbf{\checkmark}\\
            
    9 & 414.9 $\pm$165.5 & 431.4 $\pm$127.8 & 412.2 $\pm$127.9 & \text{\sffamily X} & \text{\sffamily X}\\
             
    10 & \textbf{439.5 $\pm$157.7} & 487.1 $\pm$151.0 & 512.1 $\pm$148.6 & \textbf{\checkmark} & \textbf{\checkmark}\\
           
    11 & \textbf{490.3 $\pm$181.7} & 514.2 $\pm$148.3 & 561.4 $\pm$189.3 & \text{\sffamily X}& \textbf{\checkmark}\\
              
    12 & 489.1 $\pm$207.5 & 527.3 $\pm$161.2 & 490.2 $\pm$195.5 & \text{\sffamily X} & \text{\sffamily X}\\
           
    13 & \textbf{581.8 $\pm$173.4} & 739.9 $\pm$214.4 & 714.3 $\pm$235.4 & \textbf{\checkmark} & \textbf{\checkmark}\\
             
    \bottomrule
    \multicolumn{6}{c}{$t_{tf}$ values in \textbf{bold} show significance in one way ANOVA analyses, p$<$0.05} \\
\end{tabular}
}
\vspace{-2mm}
\end{table}
\begin{figure}[b]
  \centering
  \setlength\abovecaptionskip{-0.4\baselineskip}
  \includegraphics[width=0.8\linewidth,height=4.3cm,trim={0.7cm 6.8cm 2.3cm 7.3cm},clip]{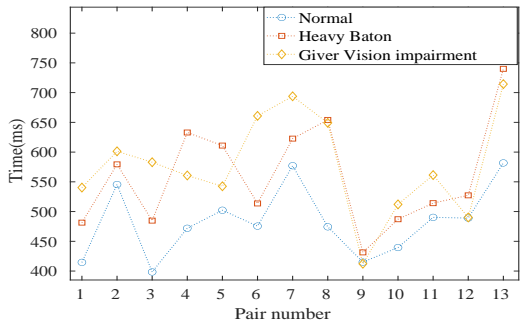}
  \caption{Mean transfer time}
  \label{fig:t_transfer}
\end{figure}

Considering the haptic interaction in a handover, the time taken to transfer the object has been defined as the time from initial contact by the taker to final release by the giver \cite{chan_grip_from_load_second_PR2}. As shown in the Fig. \ref{fig:forces_handover}, the taker's contact ($t_{tak\_con}$) was determined when the taker-$F^g$ rose above a threshold before the intersection point at $t = 0$ ms. Similarly, the giver's final release ($t_{giv\_rel}$) is determined when the giver-$F^g$ falls below the threshold. This threshold was set to 0.4 N upon investigating the sensor noise in the recorded data. The transfer time is then given by
\begin{equation}
    t_{tf}=t_{giv\_rel}-t_{tak\_con}
    \nonumber
\end{equation}
Variation in the mean transfer time for each pair and among the three settings is shown in Fig. \ref{fig:t_transfer}. It can be distinctly seen that both a rise in weight and a giver sensory impairment result in an increase in mean $t_{tf}$. For further analysis, we needed to determine the significance of the effect on $t_{tf}$ under different settings. The difference between the settings is found significant with $p < 0.0001$ for a one-way ANOVA analysis across the entire trial population with pair-number as a random effect. For each participant-pair, we also performed t-test between the recorded $t_{tf}$ for pair of settings (S1-S2, S1-S3) and one way ANOVA analysis among the three settings. A significance level of $\alpha \text{ = } 0.05$ was considered and the analysis is summarized in Table \ref{tab:transfer time} which also shows the mean $t_{tf}$ and standard deviation (SD).
As per t-test analysis, an increase in weight caused a significant increase in $t_{tf}$ for 69.2\% participants. With giver vision impairment, a significant increase in $t_{tf}$ was seen for 76.9\% participants.


\subsection{Grip release time - $t_{gr}$}

The grip release time is the total time a giver takes to release the object in handover, which we believe to be of particular importance
for the design of a robotic giver based on human data. 
This time was calculated by observing the total time in which the giver's grip force ($F^g$) decreased to zero. The giver grip release was considered to start when a first decrease in giver-$F^g$ was observed after the taker's contact, and it ended when the giver forces were reduced below the minimal threshold (at $t_{giv\_rel}$). 

\begin{table}[t]
\setlength\abovecaptionskip{-0.5\baselineskip}
\vspace{1.5mm}
  \caption{Grip Release time}
  \centering
  \label{tab:grip release time}
  \resizebox{0.9\columnwidth}{!}{
  \begin{tabular}{cccccc}
    \toprule
    
    
    Pair & \multicolumn{3}{c}{$t_{gr}$ (Mean $\pm$ SD, ms)} &     \multicolumn{2}{c}{\textit{t}-test significance}\\ \cmidrule(lr){2-4} \cmidrule(lr){5-6}
    No. & S1 & S2 & S3 & S1-S2 & S1-S3\\
    
    \midrule
    1 & \textbf{400.1 $\pm$114.3} & 475.5 $\pm$105.7 & 509.6 $\pm$153.0 &  \textbf{\checkmark} &  \textbf{\checkmark} \\
              
    2 & 522.5 $\pm$163.1 &  570.7 $\pm$152.6 &  576.6 $\pm$187.5 &  \textbf{\checkmark} & \text{\sffamily X} \\
             
    3 & \textbf{388.6 $\pm$123.6} &  469.3 $\pm$140.9&  555.2 $\pm$196.3  &  \textbf{\checkmark}  &  \textbf{\checkmark} \\
             
    4 & \textbf{437.5 $\pm$160.1} &  618.7 $\pm$169.0&  514.2 $\pm$162.8  &  \textbf{\checkmark}  &  \textbf{\checkmark} \\
          
    5 & \textbf{473.4 $\pm$154.6} &  591.7 $\pm$133.9 &  485.4 $\pm$147.2 &  \textbf{\checkmark}  & \text{\sffamily X} \\
                 
    6 & \textbf{458.6 $\pm$178.5} &  505.1 $\pm$167.6 &  632.3 $\pm$170.7&  \textbf{\checkmark}  &  \textbf{\checkmark} \\ 
               
    7 & \textbf{559.5 $\pm$155.4} &  609.6 $\pm$169.6 &  660.4 $\pm$167.2 &  \textbf{\checkmark}  &  \textbf{\checkmark} \\
                 
    8 & \textbf{454.1 $\pm$153.6} &  633.5 $\pm$150.8 &  619.4 $\pm$163.8 &  \textbf{\checkmark}  &  \textbf{\checkmark} \\
                
    9 & 397.5 $\pm$145.9 &  427.2 $\pm$121.8 &  392.5 $\pm$121.2 & \text{\sffamily X}  & \text{\sffamily X} \\
                 
    10 & \textbf{420.5 $\pm$137.1} &  480.7 $\pm$145.2 &  476.4 $\pm$133.9 &  \textbf{\checkmark}  &  \textbf{\checkmark} \\
                
    11 & \textbf{449.3 $\pm$155.5} &  498.8 $\pm$141.3 &  520.1 $\pm$175.6 &  \textbf{\checkmark} &  \textbf{\checkmark} \\
                  
    12 & 455.4 $\pm$175.8 &  514.5 $\pm$151.3 &  454.2 $\pm$173.7 &  \textbf{\checkmark}  & \text{\sffamily X} \\
               
    13 & \textbf{553.6 $\pm$153.3} &  702.9 $\pm$192.8 &  646.3 $\pm$213.0 &  \textbf{\checkmark}  &  \textbf{\checkmark} \\
                 
    \bottomrule
    \multicolumn{6}{c}{$t_{gr}$ values in \textbf{bold} show significance in one way ANOVA analyses, p$<$0.05} \\
\end{tabular}
}
\vspace{-1.0mm}
\end{table}

\begin{figure}[hb]
  \centering
  \setlength\abovecaptionskip{-0.4\baselineskip}
  \includegraphics[width=0.8\linewidth,height=4.3cm,trim={0.7cm 6.7cm 2.3cm 7.3cm},clip]{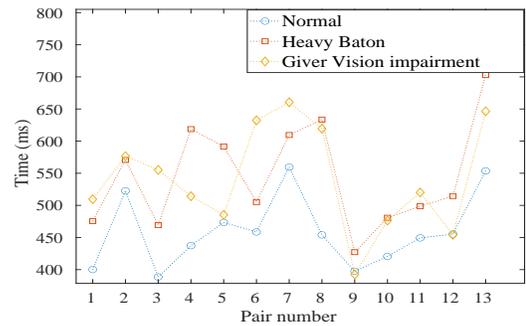}
  \caption{Mean grip release time}
  \label{fig:time_grip_release}
\end{figure}

Fig. \ref{fig:time_grip_release} shows the variation in mean $t_{gr}$ for different pairs across different experimental settings and Table \ref{tab:grip release time} summarizes the significance analysis. 
We observer that for most pairs, $t_{gr}$ increases with heavier baton and giver vision impairment. The results of a one-way ANOVA analysis with pair number as random effect (p$<$0.001) show that the difference is found significant across all participants. As per t-test analysis, an increase in weight caused a significant increase in $t_{tf}$ for about 92.3\% participants. With giver vision impairment, a significant increase in $t_{tf}$ was seen for approximately 69.2\% participants.

\subsection{Interaction Forces - Pull force}
In human-robot handovers, other than vision input, the robot can rely on interaction forces to detect, adapt, and perform the give or take in handover. For the input of interaction forces, robots are often equipped with wrist sensors or can also use end-effector torque estimation. Hence, we believe the analysis of interaction forces to be important. Multiple robot to human handovers experimentation studies used pulling force to command grip release \cite{human_human_to_human_robot_study_Controzzi,survey_review_2022_object_handovers}.

\begin{figure}[t]

    \centering
    \subfloat[]{\includegraphics[width=.45\linewidth,height=3.5cm,trim={0.1cm 5.8cm 1.5cm 6cm},clip]{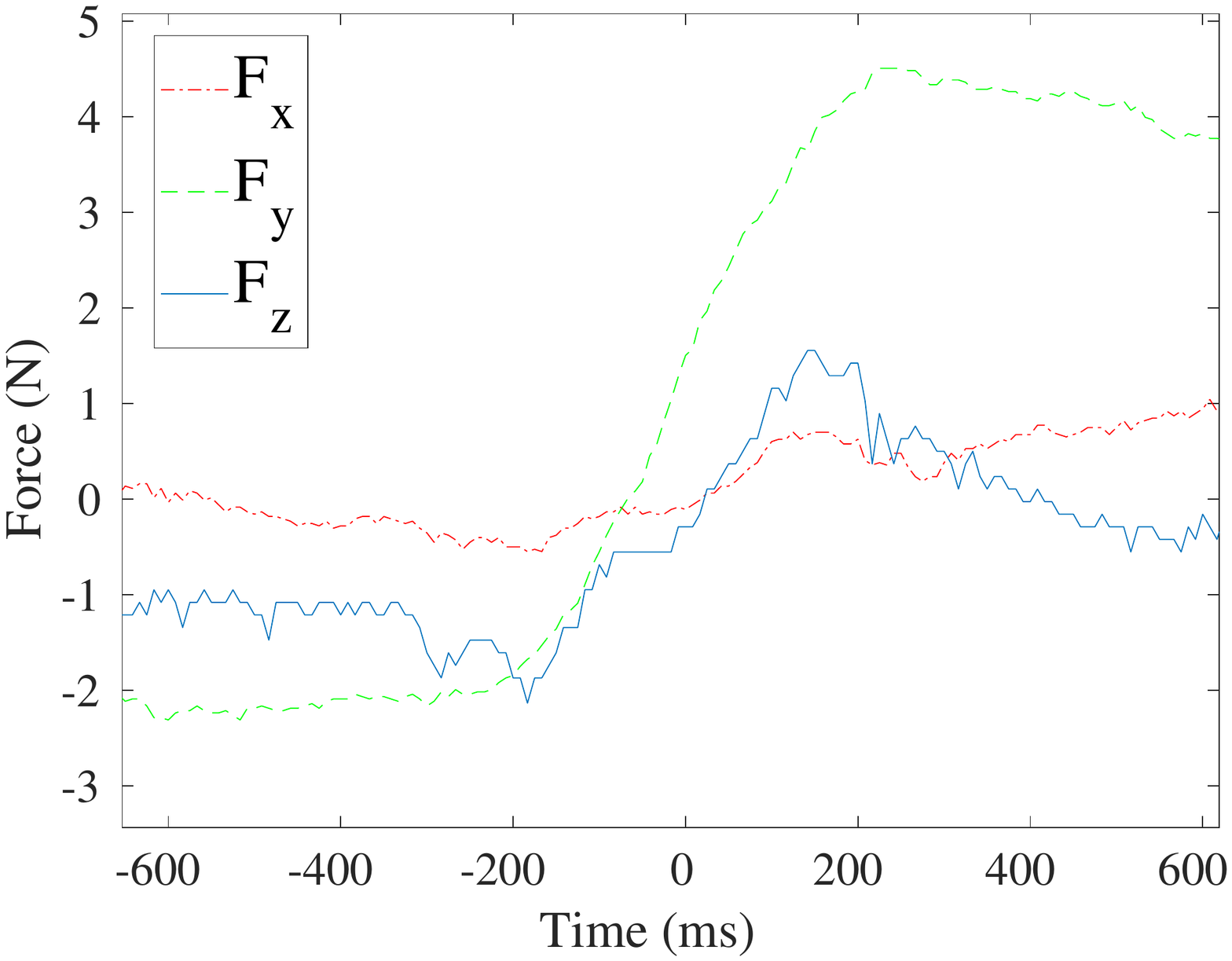}}
    \subfloat[]{\includegraphics[width=.54\linewidth,height=3.2cm,trim={{2.0cm 3.7cm 4.0cm 4.7cm}},clip]{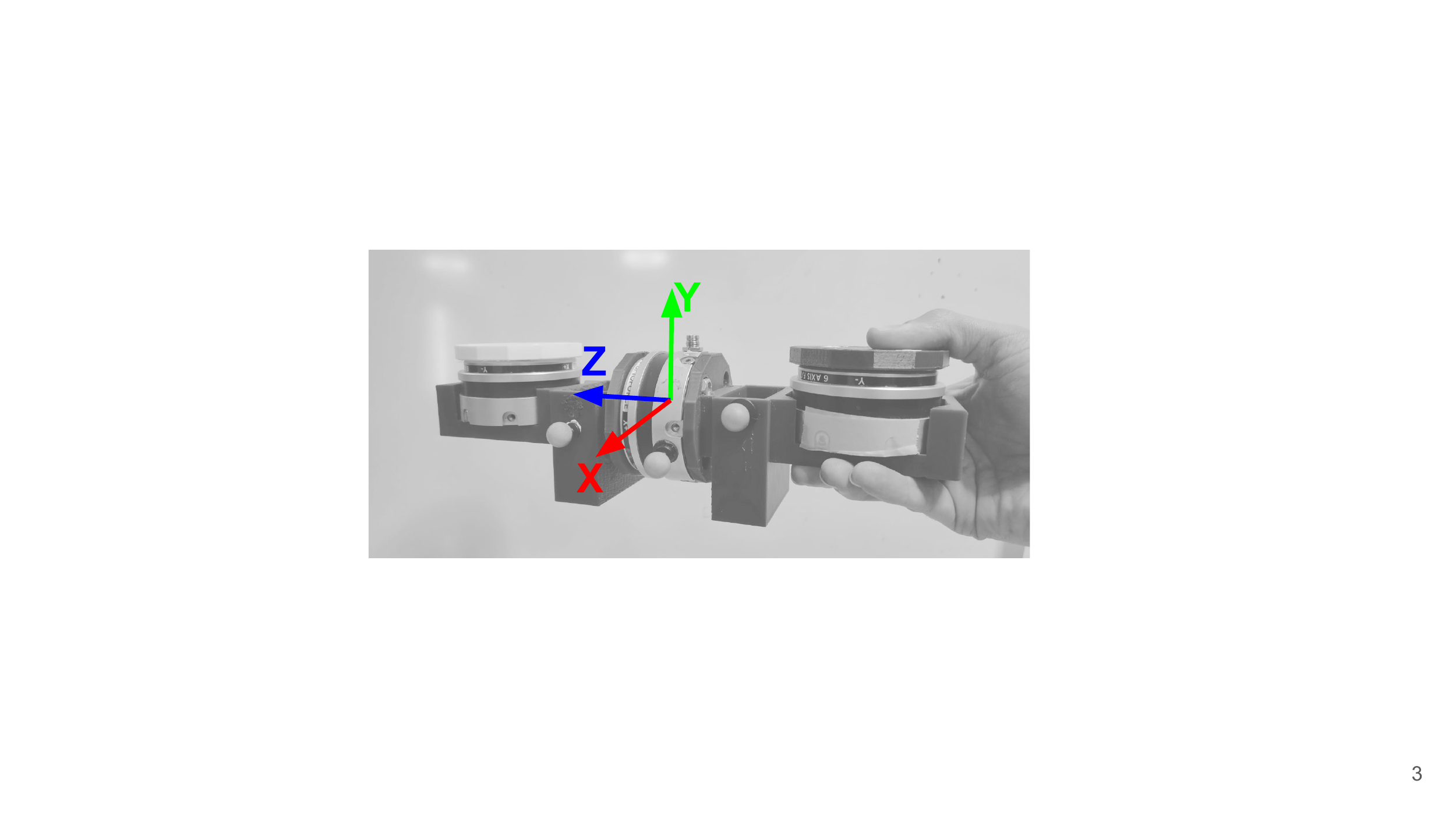}}
   
    \caption{Interaction Force: (a) Plot (b) Interaction sensor}
    \label{fig:intx_forces_handover}
\end{figure}

Considering our experimental case of horizontal transfers, the pulling force is in the direction of baton transfer. As a result, when a giver hands over the baton, the pull force in our baton is given by change in the $F_z$ component of forces measured by interaction sensor (Fig. \ref{fig:intx_forces_handover}). We analyzed the maximum pull force seen in the handovers by considering the maximum change observed in $F_z$ during the transfer time.
\begin{figure}[b]
  \centering
  \setlength\abovecaptionskip{-0.3\baselineskip}
  \includegraphics[width=0.8\linewidth,height=4.3cm,trim={1.0cm 6.5cm 2cm 7.1cm},clip]{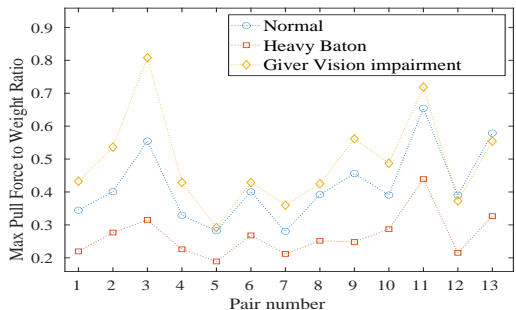}
  \caption{ Mean of maximum pull force normalized to weight}
  \label{fig:max_pull_force_normalized_for_weight}
\end{figure}
Fig. \ref{fig:max_pull_force_normalized_for_weight} shows the average maximum-pull force normalized to object weight across different settings. Across most participant pairs, we found this ratio to be higher when giver vision was impaired as compared to normal handovers. This is supported by significance in one-way ANOVA analysis with pair number as random effect (p$<$0.05).
However, a lower ratio is seen for heavier baton transfer. This led to another analysis inspecting the absolute value of maximum pulling force, whose average variation is shown in Fig. \ref{fig:max_pull_force}. We observe that the average pull force is higher for heavy baton transfer than the normal baton transfer (p$<$0.05). 
\textbf{\begin{figure}[t]
  \centering
      \setlength\abovecaptionskip{-0.3\baselineskip}      \includegraphics[width=0.8\linewidth,height=4.3cm,trim={1.0cm 6.5cm 2cm 7.1cm},clip]{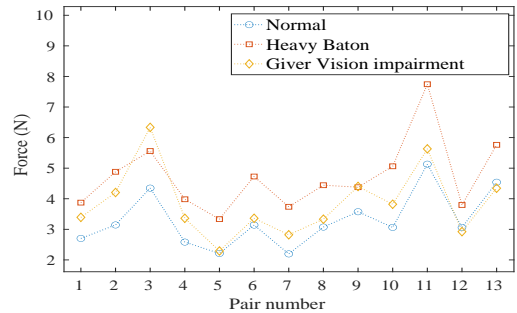}
  \caption{Mean of maximum pull force}
  \label{fig:max_pull_force}
\end{figure}}

\subsection{Interaction Forces - Load-sharing}
When humans carry an object, they feel the load (weight) of the object. This load reduces as the object is being handed over to the other person, as load-sharing between giver and taker occurs. The handover ends in complete transfer of load from giver to taker. 
In case of horizontal handovers, the felt load corresponds to the vertical component of interaction forces ($F_y$) in Fig. \ref{fig:intx_forces_handover}. Load-sharing is considered by a robotic giver in \cite{loadsharing_pull_strategy-10.3389/frobt.2021.672995,loadshare_strategy-7803296}, where the object was considered to be in a \textit{sharing state} when the measured vertical load ($F_y$) reduced to a fraction of the initial load detected ($F_y^i$) due to load-sharing by human taker. For robot to human handovers in \cite{loadsharing_pull_strategy-10.3389/frobt.2021.672995}, this fraction set to 0.5. However, the grip release was commanded by a pull force threshold detection.

An interesting analysis is to study correspondence between load-share shift/transfer to grip-force intersection. In the baton transfer, the load-share shift occurs when measured $F_y$ changes its sign, i.e. in our case changes from negative to positive, that is at time 
\begin{equation}
t_{ld\_shift} = min(t : F_y(t)>0 , t \in [t_{tak\_con},t_{giv\_rel}]) 
\nonumber
\end{equation}

\begin{figure}[b]
  \centering
  \setlength\abovecaptionskip{-0.3\baselineskip}
  \includegraphics[width=0.8\linewidth,height=4.3cm,trim={0.8cm 6.8cm 2.3cm 7.7cm},clip]{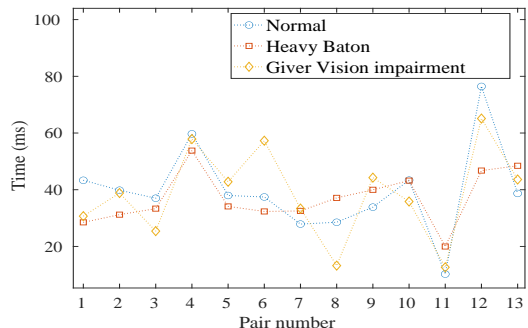}
  \caption{Mean time of load-share shift}
  \label{fig:tfull load from mid}
\end{figure}

The dominant grip force changes at the intersection for grip forces at $t=0$ ms for each handover in our data set. Hence, it is sufficient to observe absolute $t_{ld\_shift}$ values to compare the load-share shift to the intersection point of grip forces. A plot of mean $t_{ld\_shift}$ observed is given in Fig. \ref{fig:tfull load from mid}. A positive mean is observed across all settings and it is found that the values are significantly above zero (p$<$0.05 in t-test analysis) for all pairs. Thus, the load-share shift occurs after the grip forces' intersection, across all experimental settings. In other words, we validate that the dominant grip force changes before humans actually feel the load-share shift. 


\subsection{Height of Transfer}
\begin{figure}[t]
  \centering
  \setlength\abovecaptionskip{-0.1\baselineskip}
  \includegraphics[width=0.8\linewidth,height=4.3cm,trim={0.2cm 5.7cm 2cm 6.6cm},clip]{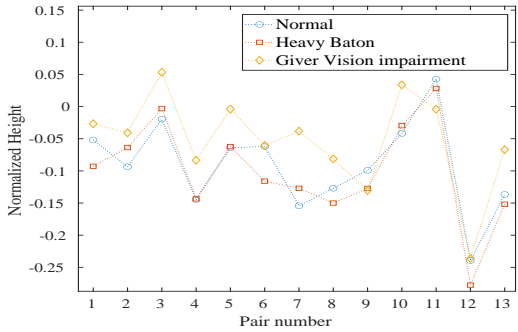}
  \caption{Mean normalized height of baton transfer}
  \label{fig:location_transfer_height}
\end{figure}
In \cite{h2H_handovers_how_dis_and_obj_mass_matter_Clint2017}, the analysis of a human-human study showed that transfer height does not depend on the object weight in handovers. 
In our analysis, for all handovers, we consider the baton transfer height as the measured \textit{Z} coordinate of baton at the intersection of grip forces, i.e. at $t=0$ ms. For proper analysis across different pairs, this was normalized to the average chest height of giver and taker. The mean variation in the normalized height across different settings is shown by Fig. \ref{fig:location_transfer_height}. A significant difference is seen in the height of transfer when comparing the three via the results of one-way ANOVA analysis with pair number as random effect (p$<$0.001). Posthoc analysis show that the height of transfer is higher for the Setting 3. We observe a higher mean height of transfer with 69.23\% participant-pairs in case with giver vision impairment. We also corroborate that the weight of the object had no significant effect on the height.

\section{Design Implications}
Based on our analysis, we propose some design implications for robots in human-robot handover scenarios.
\subsubsection{Transfer Time}
Our overall analysis shows that the $t_{tf}$ is close to 500 ms which agrees well with the reported average $t_{tf}$ for human handovers in literature \cite{survey_review_2022_object_handovers}. Our study further suggests that in human-robot handovers, a $t_{tf}$ of 500 ms shall be targeted. 
As a significant increase in weight caused increase in $t_{tf}$, a robot should expect and plan for a longer $t_{tf}$ for heavier objects or if the weight increases in the interaction. In cases when a giver is not looking, a longer $t_{tf}$ suggests that both giver and taker adapt for sensory restrictions of the giver. The robot should also adapt for a longer $t_{tf}$ if the giver is not actively looking at the handover.
\subsubsection{Grip Release time}
A higher grip release time with a heavier object for most participants is an interesting finding, as this indicates humans adapt their grip release time according to the object weight. In case of a similar, but heavy object, this might imply that more caution is exercised by humans to prevent handover failure or object fall. Thereby, in a human-robot interaction scenario, a human inspired robotic giver shall adapt its $t_{gr}$ for heavier objects. If a robot, during the interaction, detects that the current object is significantly heavier than other objects, it should increase its $t_{gr}$ for this handover. This will be especially important for robot givers relying on pre-planned grip force modulation \cite{human_human_to_human_robot_study_Controzzi}. Without vision input, the giver relies only on haptic feedback for grip release, which leads to a larger resistance by the giver before releasing, indicated by a larger $t_{gr}$. This leads to a significantly larger pull force by the taker. Thus, a robotic taker should expect more resistance in releasing the object by a human giver who is not actively looking during the handover. A robotic giver which increases its $t_{gr}$ should also expect a corresponding increase in pulling force as well. 

\subsubsection{Interaction Forces}
With the heavier baton, the increase in pull fores corresponds to increased caution displayed by both giver and taker to enable successful transfer of the heavier baton. Overall, we can say that a heavier object would lead to higher pulling forces in the handover, but the increase in pulling forces is not directly proportional to the object weight. So, humans tend to pull more while taking a heavier object but will not generally pull or expect to pull above a certain force. 
A robotic giver or taker should also expect increased resistance and pulling force with a high increase in object weight.

It is important to note that a robot giver only has the input of interaction forces and not the grip forces of a human taker. Our analysis with $F_y$ implies that it is safe for the robot to assume the taker has applied sufficient grip force when load-share shift occurs. If grip release is not yet triggered based on any other strategy, the robot can start decreasing the grip forces at the load-share shift. Such a strategy can ensure more safety in robot to human handovers, especially in cases of uncertainty like a sudden increase in the transfer object's weight that was not previously known to the human taker. However, this robotic grip-release will be commanded later than an ideal human giver's grip-release. Thus, the robot should adopt a higher grip-release time ($t_{gr}$) to preserve the naturalness by aiming for a total $t_{tf}$ of 500ms.

\subsubsection{Transfer Height}
As per our analysis, transfer height increases for a human giver when they rely on just haptic feedback and either do not or can not use the visual input. Thus, a robot taker should adapt by planning for higher transfer height when a human giver is not looking while giving the object.

\section{Learning Grip Release}
To perform a quantitative assessment of the data set, a human-inspired data-driven strategy was proposed to command robotic grip release in robot to human handovers. The creation of this strategy and further evaluation via human-robot experimentation have been discussed in detail in \cite{parag-humanoids}. During a handover, it is necessary for a robotic giver to determine when to start the grip release as the human taker forms its grip on the object. The external wrench due to human interaction can be measured by a F/T sensor placed at the robot wrist. The idea is to investigate the interaction wrench ($W_{int}$) in human handovers and then predict the start of giver grip release based on $W_{int}$. 
\subsubsection{Data-driven strategy}
For each recorded handover, $W_{int}$ comprises of a time series of 6 F/T components which vary significantly in the handover, as seen for forces in Fig. \ref{fig:intx_forces_handover}. As seen in Fig. \ref{fig:forces_handover}, the giver's grip forces starts to decline before the intersection point ($t=0$ ms). The learning task is to predict the grip release start from the observed $W_{int}$ before $t=0$ ms. Thereby, we sampled a time series of $W_{int}$ ending at a time-step $t_e$ from each saved handover:  
 \begin{equation}
 X(t_e)= \{W_{int}^{t}, t \in [t_e\text{-} 100,t_e]\}.
 \label{eqn:time_series_Int_wrench}
 \end{equation} 
 This 100 step time series corresponds to 833 ms at 120Hz for our recorded data, with $t_e$ varying between 0 to -1803 ms. Considering the average $t_{tf}$ of 500 ms \cite{chan_grip_from_load_second_PR2,survey_review_2022_object_handovers}, the labels ($l$) to the samples were assigned based on $t_e$, given by Equation \ref{eqn:Label creation te}. If the sample's $t_e$ lies after $t=-250$ ms, i.e, in the interval $(-250,0]$ ms, the input time series of $W_{int}$ corresponds to a grip release start ($l=1$). Otherwise, the sample does not represents a grip release start ($l=0$).    
 \begin{equation}
    l(X(t_e))=\begin{cases}
    1,& \text{if } t_e>-250ms\\
    0,& \text{if } t_e<=-250ms
    \end{cases}
    \label{eqn:Label creation te}
 \end{equation}
The labeled samples were then used to train a long short-term memory (LSTM) based classifier. The trained LSTM is then used for robot to human handovers where $W_{int}$ is measured. After the robot reaches the fixed final pose for handover, the time series of $W_{int}$ is continuously fed to the classifier to command the grip-release. The LSTM classifier assigns the labels 1 and 0 implying whether the input $W_{int}$ corresponded to a grip release start in human handovers.
\subsubsection{Robot to Human handovers experimentation}
The data-driven strategy was compared against loadshare and pull force based grip release strategies (Section 5.3, 5.4) in robot to human handovers. An experimental study was designed involving 20 participants taking objects from Baxter robot \cite{parag-humanoids}. The robot giver measured $W_{int}$ via a wrist F/T sensor and relied only on this sensory information to detect the handover, commanding its grip release. It was shown that the human-inspired strategy led to faster detection of handovers with the training object, reducing the $t_{tf}$. This data driven strategy was also ranked most natural by 95\% participants and preferred over other strategies with high significance. 

\section{Conclusion and Future Work}
A human study involving 26 participants in 13 pairs was done to explore multi-sensory information in human handovers. In this study, a proper experimental setup was devised to study the effect of a high increase in transfer weight and the effect of a lack of visual sensory feedback with a human giver. The data from study was processed to segregate individual handovers, leading to a multi-sensory large data set of human handovers. Further, we present a detailed analysis of the effect of the experimental settings on various aspects of human handovers. We also discuss possible design implications for a robot aiming for handovers between humans and robots. As a quantitative evaluation of the dataset, we propose and evaluate a data driven strategy for grip release of a robotic giver in robot to human handover.

In the future, we intend to investigate how we can make better use of our human handover dataset in a quantitative setting. We plan to further research on applying our work to produce more realistic human-robot handovers and what human-inspired metrics may be applied to assess their quality. It will also be interesting to study how human perceptions of a natural handover change when a robot is involved.
\section*{ACKNOWLEDGMENT}
This work was supported by Digital Futures at KTH. 
\vspace{0.1mm}
\vspace{-2mm}
\bibliographystyle{IEEEtran}
\bibliography{IEEEabrv,pk_ref_paper2_datasetHRI.bib}
\end{document}